\documentclass[runningheads]{llncs}
\usepackage{graphicx}
\usepackage[utf8]{inputenc} 
\usepackage[T1]{fontenc}    
\usepackage{hyperref}       
\usepackage{url}          
\usepackage{booktabs}       
\usepackage{amsmath, amssymb, amsfonts}
\usepackage{nicefrac}      
\usepackage{microtype}     
\usepackage{xcolor} 
\usepackage{multirow}
\usepackage{array}
\usepackage{makecell}
\hypersetup{
    colorlinks=true,
    urlcolor=blue
}
\begin{document}

\title{ESRVS: Extreme Semi-Supervised Retinal Vessel Segmentation with a Single Annotated Image}
\titlerunning{ESRVS: Single-Annotation Semi-Supervised Segmentation}
\author{
Mingzhi Xu\inst{1} \and
Yizhe Zhang\inst{1}\thanks{Corresponding author}
}
\institute{
\textsuperscript{1}Nanjing University of Science and Technology, Nanjing, China\\
\email{zhangyizhe@njust.edu.cn}
}

%
\maketitle              
\begin{abstract}
Learning from minimal human supervision is a long-standing goal in medical image analysis, where dense expert annotations are costly. We study retinal vessel segmentation in an extreme semi-supervised setting with one annotated image and a pool of unlabeled images. We propose ESRVS, which selects a representative reference image for manual annotation and transfers vessel cues using target-domain-adapted DINOv3 features. ESRVS constructs a multi-granular vessel prototype, combines prototype-similarity maps with a physics-inspired prior to generate initial pseudo-labels, and refines the transferred supervision through weighted pseudo-label training and adversarial refinement. Across eight public datasets, ESRVS achieves the best Dice and clDice on six datasets, and the best HD95 on all eight datasets among the compared semi-supervised methods, although those methods use 10–20\% labeled data. With Mask2Former, ESRVS retains on average 93.7\% of fully supervised Dice and 95.1\% of fully supervised clDice. These results demonstrate the potential of foundation-model label propagation for highly label-efficient retinal vessel segmentation. Code is available at \url{https://github.com/IAANNH/ESRVS}.

\end{abstract}

\section{Introduction}
Retinal vessels provide a critical window into both ocular and systemic health, serving as indispensable biomarkers for diagnosing conditions such as diabetic retinopathy and glaucoma~\cite{kellner2024eye}. Consequently, the accurate segmentation of these vascular networks is essential for modern AI-assisted ophthalmological diagnosis~\cite{qin2024review}. However, manual delineation of retinal microvasculature is notoriously labor-intensive, requiring specialized clinical expertise. Furthermore, severe inter- and intra-observer variability can compromise the consistency of subsequent diagnoses. While recent advances in deep learning have achieved near-expert segmentation performance~\cite{yao2024cnn}, these models heavily rely on fully supervised training with densely annotated datasets~\cite{zhou2018unet++}. The acquisition of such large-scale, pixel-perfect medical data is prohibitively expensive and time-consuming, driving the urgent need for label-efficient segmentation paradigms.

Semi-supervised learning leverages a small set of labeled images alongside a larger pool of unlabeled data~\cite{lu2025semi}. SSL methods~\cite{lu2025semi} reduce annotation dependency by enforcing consistency or generating pseudo-labels on unlabeled images; yet, they typically still require a non-trivial proportion of labeled data (e.g., 10\% to 20\%) to prevent confirmation bias. Meanwhile, few-shot learning (FSL) approaches~\cite{alsaleh2024few} attempt to generalize from minimal examples, but often rely on complex meta-training episodes and struggle to capture the intricate, continuous topologies of dense vascular networks.

In parallel, large-scale foundation models, such as Vision Transformers (ViTs)\cite{dosovitskiy2020image} and recent DINO variants\cite{caron2021emerging,oquab2023dinov2,simeoni2025dinov3}, have demonstrated extraordinary zero-shot feature extraction capabilities. Unfortunately, their direct application in medical imaging is severely hindered by a significant domain gap: features learned from natural scenes often fail to localize the delicate and fragile structures of retinal vessels. Although some recent techniques attempt to synthesize pseudo-labels using AI-based priors~\cite{zhang2025towards}, the generated labels frequently suffer from topological fragmentation and severe background noise, severely degrading the performance of the downstream segmentation models.

Driven by the goal of extreme annotation efficiency, recent studies have explored one-label semi-supervised medical image segmentation~\cite{zhang2026medical}. However, retinal vessel segmentation remains challenging under such limited supervision due to the complex vascular topology, diverse vessel morphology, and the need to preserve fine structures. To address this challenge, we propose \textbf{ESRVS (Extreme Semi-supervised Retinal Vessel Segmentation)}, a framework for label-efficient retinal vessel segmentation (see Figure~\ref{fig1}).
ESRVS leverages a single annotated reference image and its corresponding label, along with a large pool of unlabeled images. Multi-granular features are extracted from the labeled pair using a DINOv3 foundation model enhanced with task-specific self-supervision. These features are then refined with adaptive physics-inspired priors to generate pseudo-labels, which guide a shape-aware network in correcting topological errors. Our main contributions are threefold:
\textbf{(1)} We propose \textbf{ESRVS}, a retinal vessel segmentation framework that pushes label efficiency to the extreme by learning from exactly one annotated image, bridging the gap to fully supervised performance.
\textbf{(2)} We design a novel propagation mechanism that integrates DINOv3-based multi-granular features with physics-inspired priors. This generates high-fidelity pseudo-labels from the single reference to guide continuous model refinement.
\textbf{(3)} Extensive experiments across eight public datasets demonstrate our method's exceptional data efficiency. Using only one labeled image, ESRVS with Mask2Former retains over 93\% of the fully supervised Dice and clDice performance on seven of the eight datasets, excluding FIVES.

\section{Related Work}

\subsection{Fully Supervised learning}
Fully supervised learning remains the cornerstone of retinal vessel segmentation, with U-Net~\cite{ronneberger2015u} and its derivatives serving as the predominant backbones. To tackle the intricate microvasculature, subsequent research has focused on enhanced feature extraction~\cite{gu2019net}, multi-scale feature fusion~\cite{liu2023wave}, and context-aware encoding~\cite{wu2021scs}. More recently, Transformer-based architectures have gained traction for modeling global context and long-range dependencies, ranging from hybrid CNN-Transformer networks~\cite{jiang2024covi} to pure transformer-based models~\cite{tan2023oct2former}. While these methods achieve excellent segmentation fidelity, they are inherently constrained by the requirement for massive, pixel-perfect annotations, rendering them highly sensitive to annotation scarcity and difficult to scale in clinical deployment.

\subsection{Semi-Supervised Learning}
Semi-supervised learning (SSL) aims to alleviate annotation costs by leveraging unlabeled data alongside a sparse set of annotations. The prevailing strategies generally fall into two categories: pseudo-labeling and consistency regularization. Pseudo-labeling generates temporary targets for unlabeled data, often refined through confidence filtering, probability map sharpening~\cite{moezzi2023uncertainty}, and intra-batch entropy minimization~\cite{zheng2025crossnext}. For example, CrossNeXt~\cite{zheng2025crossnext} combines intra-batch entropy minimization with cross-teaching between networks to refine predictions and enhance segmentation accuracy. Meanwhile, consistency regularization enforces prediction stability against input noise or feature-level perturbations~\cite{ouali2020semi}, and has recently been fortified by teacher-student distillation~\cite{zhao2024semi} and adversarial training constraints~\cite{you2022fmwdct}. Despite their success, SSL paradigms still implicitly assume the presence of a meaningful fraction of labeled data (typically 10-20\%), causing them to suffer from catastrophic confirmation bias when labels are reduced to an extreme single-label regime.

\subsection{Few-Shot and One-Shot Learning}
To address extreme label scarcity, few-shot and one-shot segmentation have emerged as efficient alternatives. Most existing methods adopt meta-learning paradigms~\cite{jing2020self} to leverage transferable knowledge from known to unseen classes. A dominant strategy is prototype-based learning, which classifies query samples by measuring their embedding similarity to learned class prototypes~\cite{wang2019panet}. However, for complex anatomical structures like retinal vessels, single-level global prototypes are fundamentally insufficient to capture drastic structural and topological variations. To alleviate this, recent works have incorporated self-supervised (SS) to exploit latent structural priors, utilizing superpixel-based frameworks~\cite{ouyang2020self} or multi-scale, multi-region prototypes~\cite{sun2026multi} to account for shape variations. Despite these advancements, prototypical representations inherently compress spatial details and underutilize the pixel-level relationships in unlabeled data, leading to severe topological disconnections when training data is reduced to a single annotated sample.

\begin{figure}[t]
\centering
\includegraphics[width=1\textwidth]{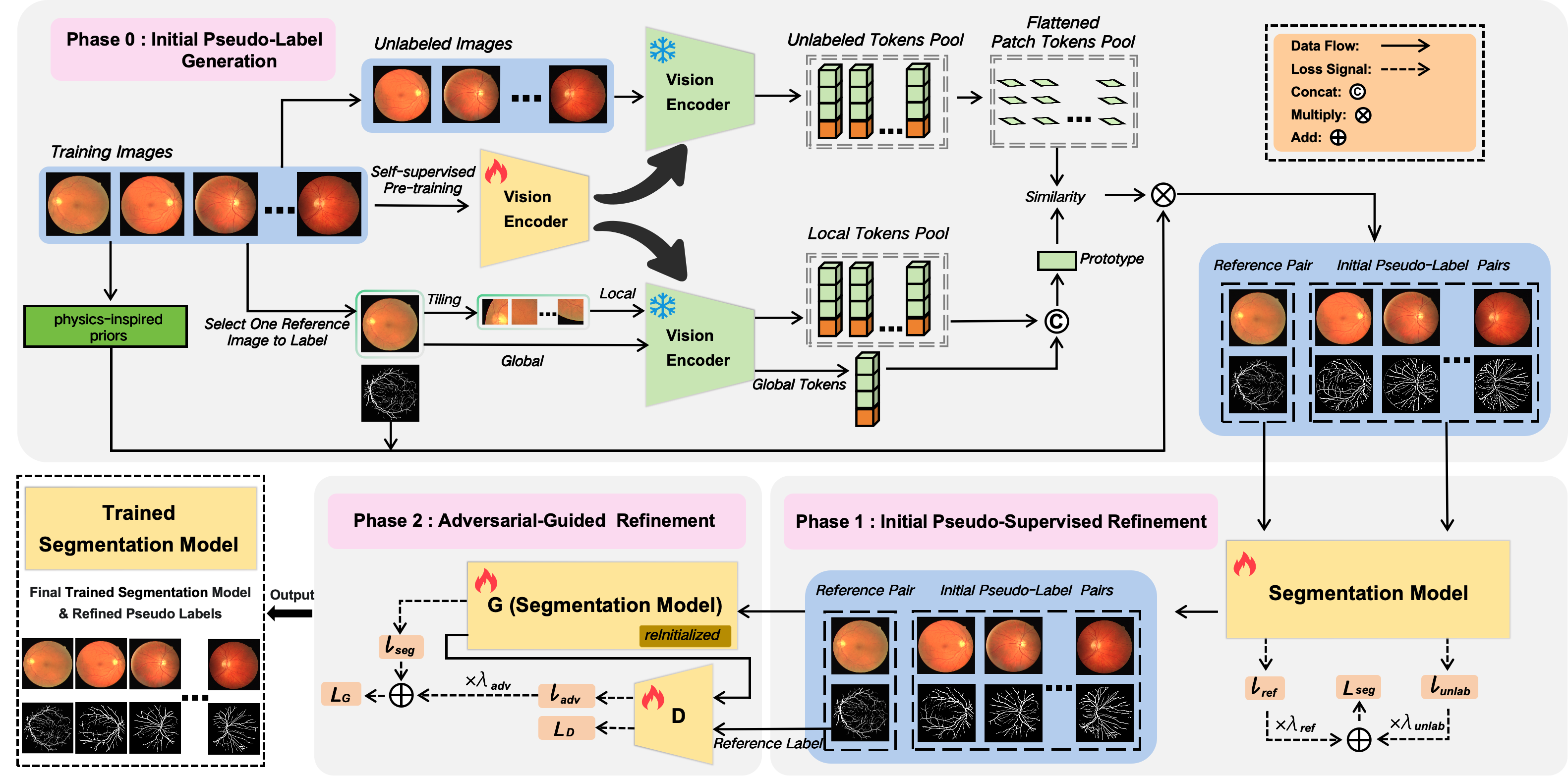}
\caption{The ESRVS framework for retinal vessel segmentation.} 
\label{fig1}
\end{figure}


\section{Method}
The ESRVS framework leverages a self-supervised foundation model (DINOv3) to propagate supervision from a single annotated reference image to an unlabeled dataset. As illustrated in Figure~\ref{fig1}, the pipeline consists of three phases: (1) Initial Pseudo-Label Generation; (2) Initial Pseudo-Supervised Refinement; and (3) Adversarial-Guided Refinement for segmentation enhancement.
\subsection{Problem Setup}
Let $\mathcal{D} = \{I_i\}_{i=1}^N$ denote a fundus image dataset containing $N$ samples, where each image $I_i \in \mathbb{R}^{H \times W \times 3}$. 
Our goal is to train a segmentation model $\mathcal{S}_\phi$ (parameterized by $\phi$) using only a single annotated reference image $x_{ref} = (I_{ref}, L_{ref})$ and a set of unlabeled images.
Here, $I_{ref}$ denotes the reference fundus image and $L_{ref} \in \{0, 1\}^{H \times W}$ is its binary vessel label. The remaining images constitute the unlabeled set, denoted as $\mathcal{D}_{u} = \mathcal{D} \setminus \{I_{ref}\}$. Throughout the paper, $\mathcal{D}$ denotes the training split only. Reference selection, target-domain self-supervised adaptation, pseudo-label generation, and model training are restricted to the training images; held-out test images and their annotations are used only for final evaluation.

\subsection{Initial Pseudo-Label Generation}
\noindent \textbf{Representative Reference Selection.} We first select the reference image using image features only, without accessing any segmentation annotations. We then simulate annotation by revealing the ground-truth mask of the selected image; all remaining segmentation masks are hidden during training. To obtain a representative supervision signal, we adopt a centroid-based reference selection strategy. Specifically, a pre-trained feature encoder $\mathcal{E}_{R}$ (e.g., DINOv3) embeds all images in $\mathcal{D}$ into a feature space. We compute the dataset centroid $C$ as the mean feature embedding:
\begin{equation}
C = \frac{1}{|\mathcal{D}|}
\sum_{I_i \in \mathcal{D}} \mathcal{E}_{R}(I_i).
\end{equation}
The reference image $I_{\text{ref}}$ is then selected as the one whose feature representation is most similar to the centroid:
\begin{equation}
I_{ref} = \arg\max_{I_i \in \mathcal{D}} 
\text{CosineSim}(\mathcal{E}_{R}(I_i), C).
\end{equation}

\noindent \textbf{Structure-Aware Semantic Amplification.} Given the selected reference $x_{ref}$, we aim to generate high-fidelity pseudo-labels for the unlabeled set $\mathcal{D}_{u}$. Direct feature matching often suffers from scale ambiguity and coarse boundary alignment. To address this, we introduce a label generation mechanism consisting of Multi-Granular Semantic Matching (MGSM) and Adaptive Physics-Inspired Priors Alignment (APPA).

\textbf{Multi-Granular Semantic Matching (MGSM).}
To achieve robust vessel segmentation across scales, we design a multi-granular matching strategy that constructs a unified vessel prototype and performs dual-view prototype matching.
 
\textit{Multi-Granular Prototype Construction.}
Given a reference image $I_{ref}$, we extract vessel-related features at two complementary granularities. At the macro level, $I_{ref}$ is processed holistically to capture global vascular topology. At the micro level, we partition $I_{ref}$ into a grid of local patches to encode fine-grained structural details. Both views are encoded using a DINOv3 backbone $\mathcal{E}_{D}$, which is further fine-tuned in a self-supervised manner on the target domain. Let $\mathcal{F}_{macro}$ and $\mathcal{F}_{micro}$ denote the sets of patch embeddings extracted from large-scale and fine-grained patches of the same image, respectively. To ensure semantic consistency, we filter the extracted embeddings using the reference annotation $L_{ref}$ and retain only vessel-relevant features: $\Omega = \left\{ \mathbf{f} \in (\mathcal{F}_{macro} \cup \mathcal{F}_{micro}) \mid L_{ref}(\mathbf{f}) = 1 \right\},$ where $\mathbf{f}$ denotes the feature vector at a spatial location with $1$ indicating the target (foreground) class. We then aggregate all valid features to form a unified vessel prototype:
\begin{equation}
\mathcal{P}_{uni} = \text{Normalize}\left( \frac{1}{|\Omega|} \sum_{\mathbf{f} \in \Omega} \mathbf{f} \right),
\end{equation}

\textit{Multi-Granular Prototype Matching.} 
Given the unlabeled target set $\mathcal{D}_{u} = \mathcal{D} \setminus \{I_{ref}\}$, for each unlabeled image $I_u \in \mathcal{D}_{u}$, we extract patch embeddings $\mathcal{F}(I_u)$ using the same encoder $\mathcal{E}_{D}$. We compute a cosine similarity map between each patch feature and the unified prototype $\mathcal{P}_{\text{uni}}$:
\begin{equation}
\mathcal{S}_{\text{fused}}(I_u) = \text{CosSim}\big(\mathcal{F}(I_u), \mathcal{P}_{\text{uni}}\big).
\end{equation}
which highlights vascular regions and serves as a coarse localization map.

\textbf{Adaptive Physics-Inspired Priors Alignment (APPA).}  
To refine coarse vascular predictions and suppress non-vascular noise in $\mathcal{S}_{fused}$, APPA incorporates an intensity-guided structural prior derived from the image domain. 

\textit{Physics-Inspired Priors Integration.}  
Retinal vessels typically appear darker in the green channel because of stronger hemoglobin absorptio~\cite{ricci2007retinal}. After normalizing $I_g$ to [0,1], we therefore define $V_{phys} = \text{Normalize}(1 - I_g).$ The refined map is obtained via:
\begin{equation}
\mathcal{M}_{sem} = \mathcal{S}_{fused} \odot V_{phys},
\end{equation}
which suppresses background noise and enhances vessel structures. 

\textit{Label Generation.}
Binary pseudo-labels $\hat{L}_{u}^{(0)}$ are obtained via adaptive local thresholding on $\mathcal{M}_{sem}$, :
\begin{equation}
\hat{L}_{u}^{(0)}(x,y) = \mathbb{I}\Big( \mathcal{M}_{sem}(x,y) > \mu_{\mathcal{N}}(x,y) + \delta \Big),
\end{equation}
where $\mu_{\mathcal{N}}$ is the adaptive local mean and $\delta$ (+0.02 by default) is a sensitivity margin. This yields a pseudo-labeled set ${x}_{u} = \{(I_u, \hat{L}_{u}^{(0)}) \mid I_u \in \mathcal{D}_{u}\}.$

\subsection{Initial Pseudo-Supervised Refinement}
We train $\mathcal{S}_\phi$ on $\mathcal{D}_{train}^{(1)} = \{(I_{ref}, L_{ref})\} \cup \{(I_u, \hat{L}_{u}^{(0)}) \mid I_u \in \mathcal{D}_{u}\},$ and minimize the following loss:
\begin{equation}
\mathcal{L}^{(1)} =
\lambda_{ref}\,\ell\big(\mathcal{S}_\phi(I_{ref}), L_{ref}\big)
+
\lambda_{u}\,\ell\big(\mathcal{S}_\phi(I_u), \hat{L}_{u}^{(0)}\big),
\end{equation}
where the Ref.Weighting $\lambda_{ref} > \lambda_{u}$ (we set $\lambda_{ref}=3$ and $\lambda_{u}=1$) to prioritize the reference supervision. The loss function is a weighted combination of cross-entropy loss and Dice loss, with equal weights of 0.5 assigned to each component.

The predictions in this phase, $\hat{L}_{u}^{(1)}$, are then selected as pseudo-labels for Adversarial-Guided Refinement and used to construct the second-stage training dataset $\mathcal{D}_{train}^{(2)} = \{(I_{ref}, L_{ref})\} \cup \{(I_u, \hat{L}_{u}^{(1)}) \mid I_u \in \mathcal{D}_{u}\}.$

\subsection{Adversarial-Guided Refinement for segmentation enhancement}
We re-initialize the segmentation network $\mathcal{S}_\phi$ as the generator $\mathcal{G}$ and train it from scratch on $\mathcal{D}_{train}^{(2)}.$ A PatchGAN-style discriminator $\mathcal{D}$ is added to enforce structural and stylistic consistency. The loss is:
\begin{equation}
\begin{aligned}
\mathcal{L}_\mathcal{G}^{(2)} =
  \lambda_{ref}\, \ell\big(\mathcal{S}_\phi(I_{ref}), L_{ref}\big) 
 + \lambda_{u}\, \ell\big(\mathcal{S}_\phi(I_u), \hat{L}_{u}^{(1)}\big)
 + \lambda_{adv}\ell_{adv}\big(\mathcal{D}                  (\mathcal{S}_\phi(I_{u})), 1\big)
\end{aligned}
\end{equation}
where $\lambda_{ref}=3$, $\lambda_{u}=1$, $\lambda_{adv}=0.1$ and $\ell_{adv}(\cdot,\cdot)$ is a binary cross-entropy for real and fake labels.
\begin{equation}
\begin{aligned}
\mathcal{L}_\mathcal{D}^{(2)} =
& \; \ell_{adv}\big(\mathcal{D}(L_{ref}), 1\big) + \ell_{adv}\big(\mathcal{D}(\mathcal{S}_\phi(I_u)), 0\big)
\end{aligned}
\end{equation}
The synergy between the generator loss $\mathcal{L}_\mathcal{G}^{(2)}$ and the discriminator loss $\mathcal{L}_\mathcal{D}^{(2)}$ ensures the network learns to:
(1) trust the supervision from the high-quality reference image, (2) correct errors in the pseudo-labels, and (3) generate outputs that are structurally and stylistically plausible.

\section{Experiments}
For a comprehensive evaluation, we compare ESRVS against a wide range of representative methods, including(1) semi-supervised segmentation methods;(2) fully supervised segmentation methods. All experiments are conducted on a machine equipped with an NVIDIA L20 GPU and 256 GB of RAM.

\noindent \textbf{Datasets.} We evaluate our method on eight public fundus vessel segmentation datasets. Specifically, the datasets include DRIVE~\cite{staal2004ridge}(40 images, classic benchmark), STARE~\cite{hoover2000locating}(20 images, some pathological cases), CHASEDB1~\cite{owen2011retinal}(28 images from a child health study), HRF~\cite{budai2013robust}(45 high-resolution images of healthy, diabetic retinopathy, and glaucoma patients), FIVES~\cite{jin2022fives}(800 images with large intra-dataset diversity), and three additional specialized datasets: ORVS~\cite{sarhan2021transfer}, 
DRHAGIS~\cite{holm2017dr}, and AVRDB~\cite{akbar2017avrdb}, covering diverse imaging modalities and clinical focuses. The specific
dataset splits are summarized in Table~\ref{tab:data_splits}. For \textbf{ESRVS}, we strictly followed a \textit{single-label} setting, using only one image with its corresponding ground truth as the labeled reference pair; all other images were treated as unlabeled. \textbf{For semi-supervised comparison methods}, a more lenient setting was adopted due to the inherent difficulty of vessel segmentation and the relatively small number of images in most datasets: for the large-scale FIVES dataset, 10\% of the training data was used as labeled, while for all other smaller datasets, 20\% of the training data was used as labeled, with the remaining images treated as unlabeled. 

\begin{table}[h]
\footnotesize
\centering
\caption{Data splits under different training settings for each dataset. All methods are evaluated on the same test sets.}
\label{tab:data_splits}
\resizebox{0.95\textwidth}{!}{
\begin{tabular}{l cc | cc | cc | cc}
\toprule

& & & \multicolumn{2}{c}{\textbf{Fully-supervised}} 
& \multicolumn{2}{c}{\textbf{Semi-supervised}} 
& \multicolumn{2}{c}{\textbf{Ours}} \\
\cmidrule(lr){4-5} \cmidrule(lr){6-7} \cmidrule(lr){8-9}

\textbf{Dataset} 
& \textbf{Total} & \textbf{Training}
& \textbf{Labeled} & \textbf{Unlabeled}
& \textbf{Labeled} & \textbf{Unlabeled}
& \textbf{Labeled} & \textbf{Unlabeled} \\
\midrule

DRIVE & 40 & 28 & 28 & 0 & 6 & 22 & 1 & 27 \\
STARE & 20  & 14 & 14 & 0 & 3 & 11 & 1 & 13  \\
CHASEDB1 & 28 & 20& 20 & 0 & 4 & 16 & 1 & 19 \\
HRF & 45 & 32 & 32 & 0 & 6 & 26 & 1 & 31 \\
FIVES & 800& 600 & 600 & 0 & 60 & 540 & 1 & 599 \\
ORVS & 49 & 34 & 34 & 0 & 7 & 27 & 1 & 33 \\
DRHAGIS & 40 & 28 & 28 & 0 & 6 & 22 & 1 & 27 \\
AVRDB & 100 & 75& 75 & 0 & 15 & 60 & 1 & 74 \\

\bottomrule
\end{tabular}
}
\end{table}

\begin{table}[t]
\centering
\caption{Performance comparison across eight public datasets. $\uparrow/\downarrow$ indicate that higher/lower is better. Our method uses only one labeled image, while semi-supervised baselines use 10\% labeled data on FIVES and 20\% on the other datasets. All methods adopt the same U-Net architecture as the segmentation backbone.}
\label{tab:unet_vs_semi}
\resizebox{0.95\textwidth}{!}{%
\begin{tabular}{cl||c||cccccc}
\toprule
& & \textbf{Single-Label} & \multicolumn{6}{c}{\textbf{Semi-Supervised Methods}} \\
\cmidrule(lr){3-3} \cmidrule(lr){4-9}
\textbf{Dataset} & \textbf{Metric} & Ours & MT & ICT & CPS & CTBCT & URPC & KnowSAM \\
\midrule

\multirow{3}{*}{\textbf{DRIVE}}
& \textbf{Dice $\uparrow$}      & \textbf{\underline{0.7102}} & 0.3194 & 0.4009 & 0.3301 & 0.3776 & 0.3086 & 0.6854 \\
& \textbf{clDice $\uparrow$}   & \textbf{\underline{0.6881}} & 0.2932 & 0.3432 & 0.3437 & 0.3444 & 0.2728 & 0.6756 \\
& \textbf{HD95 $\downarrow$}    & \textbf{\underline{9.55}}   & 19.29 & 17.91 & 65.07 & 16.43 & 63.32 & 19.53 \\
\midrule

\multirow{3}{*}{\textbf{FIVES}}
& \textbf{Dice $\uparrow$}      & 0.6668 & 0.2977 & 0.2141 & 0.4954 & 0.4235 & 0.7048 & \textbf{\underline{0.7111}} \\
& \textbf{clDice $\uparrow$}   & 0.6771 & 0.3313 & 0.2339 & 0.6029 & 0.4728 & \textbf{\underline{0.7947}} & 0.7291 \\
& \textbf{HD95 $\downarrow$}    & \textbf{\underline{64.56}} & 303.82 & 290.12 & 204.33 & 437.26 & 91.46 & 100.92 \\
\midrule

\multirow{3}{*}{\textbf{ORVS}}
& \textbf{Dice $\uparrow$}      & \textbf{\underline{0.5415}} & 0.2909 & 0.4283 & 0.1824 & 0.1761 & 0.2532 & 0.4548 \\
& \textbf{clDice $\uparrow$}   & \textbf{\underline{0.6626}} & 0.2791 & 0.4035 & 0.1918 & 0.1442 & 0.1966 & 0.3363 \\
& \textbf{HD95 $\downarrow$}    & \textbf{\underline{35.06}} & 140.48 & 125.94 & 274.62 & 217.98 & 305.66 & 271.13 \\
\midrule

\multirow{3}{*}{\textbf{STARE}}
& \textbf{Dice $\uparrow$}      & 0.6357 & 0.3336 & 0.4363 & 0.2028 & 0.1866 & 0.2979 & \textbf{\underline{0.6547}} \\
& \textbf{clDice $\uparrow$}   & \textbf{\underline{0.7283}} & 0.3126 & 0.3791 & 0.1879 & 0.1316 & 0.3009 & 0.6351 \\
& \textbf{HD95 $\downarrow$}    & \textbf{\underline{19.58}}   & 28.51 & 24.27 & 49.84 & 102.07 & 82.27 & 29.54 \\
\midrule


\multirow{3}{*}{\textbf{HRF}}
& \textbf{Dice $\uparrow$}      & \textbf{\underline{0.6347}} & 0.5042 & 0.5751 & 0.2029 & 0.3684 & 0.3682 & 0.5564 \\
& \textbf{clDice $\uparrow$}   & \textbf{\underline{0.6444}} & 0.5018 & 0.6347 & 0.2532 & 0.4296 & 0.2682 & 0.4548 \\
& \textbf{HD95 $\downarrow$}   & \textbf{\underline{44.34}} & 72.32 & 55.11 & 508.65 & 599.13 & 358.89 & 195.92 \\
\midrule

\multirow{3}{*}{\textbf{CHASEDB1}}
& \textbf{Dice $\uparrow$}      & \textbf{\underline{0.6904}} & 0.4305 & 0.4369 & 0.3086 & 0.2336 & 0.3967 & 0.6321 \\
& \textbf{clDice $\uparrow$}   & \textbf{\underline{0.6653}} & 0.3474 & 0.3678 & 0.3495 & 0.2048 & 0.3615 & 0.6279 \\
& \textbf{HD95 $\downarrow$}    & \textbf{\underline{14.78}} & 69.27 & 43.63 & 63.39 & 91.91 & 107.70 & 50.23 \\
\midrule

\multirow{3}{*}{\textbf{DRHAGIS}}
& \textbf{Dice $\uparrow$}      & \textbf{\underline{0.5841}} & 0.2681 & 0.2058 & 0.1952 & 0.2116 & 0.2133 & 0.5112 \\
& \textbf{clDice $\uparrow$}   & \textbf{\underline{0.6256}} & 0.2561 & 0.1833 & 0.2718 & 0.2025 & 0.1507 & 0.4759 \\
& \textbf{HD95 $\downarrow$}    & \textbf{\underline{102.07}} & 583.47 & 685.66 & 585.65 & 571.35 & 617.87 & 217.52 \\
\midrule

\multirow{3}{*}{\textbf{AVRDB}}
& \textbf{Dice $\uparrow$}      & \textbf{\underline{0.6387}} & 0.5957 & 0.5656 & 0.6235 & 0.2386 & 0.4468 & 0.4241 \\
& \textbf{clDice $\uparrow$}   & 0.6891 & 0.6419 & 0.6101 & \textbf{\underline{0.7542}} & 0.3108 & 0.4407 & 0.3991 \\
& \textbf{HD95 $\downarrow$}    & \textbf{\underline{24.18}} & 47.49 & 48.66 & 25.60 & 193.01 & 66.55 & 113.35 \\
\bottomrule
\end{tabular}%
}
\end{table}

\noindent
\textbf{Implementation Details.} The proposed framework is implemented in PyTorch with a multi-stage pipeline. Different input resolutions are adopted for different stages: $2048 \times 2048$ for DINOv3-based fine-grained feature extraction and prototype matching, and $1024 \times 1024$ for segmentation network training and inference. The training batch size is set to 1 pair per iteration (consisting of exactly one reference sample
and one training sample) to satisfy the one-shot constraint, while a batch size of 4 is used for offline pseudo-label inference. Segmentation performance is evaluated using Dice~\cite{dice1945measures}, clDice~\cite{shit2021cldice}, and HD95~\cite{taha2015metrics}, measuring region overlap, topology preservation, and boundary accuracy, respectively.

\noindent \textbf{Compared Methods.} We compare ESRVS with representative semi-supervised segmentation methods reflecting recent mainstream design paradigms. Consistency regularization-based methods, including Mean Teacher (MT)~\cite{tarvainen2017mean}, Interpolation Consistency Training (ICT)~\cite{verma2022interpolation}, Cross Pseudo Supervision (CPS)~\cite{chen2021semi}, and Cross Teaching between CNN and Transformer (CTBCT)~\cite{luo2022semi}, enforce prediction consistency under perturbations, interpolations, or cross-model supervision. Uncertainty-based approaches such as URPC~\cite{luo2021efficient} further refine pseudo labels by explicitly modeling prediction uncertainty. In addition, foundation model-based methods like KnowSAM~\cite{huang2025learnable} distill knowledge from Segment Anything Model (SAM) to improve semi-supervised learning. Notably, these methods typically rely on multiple annotated images, whereas ESRVS operates in a more challenging extreme setting with only a single labeled image. For fully supervised learning, we evaluate CNN-based UNet variants, a hybrid CNN-Transformer model (HSNet)~\cite{zhang2022hsnet}, and a transformer-based architecture (Mask2Former)~\cite{cheng2022masked}.

\subsection{Performance Comparison with Semi-Supervised Methods}
Table~\ref{tab:unet_vs_semi} demonstrates the extreme label efficiency of our proposed method. Using only a single annotated image, our framework outperforms most rival semi-supervised baselines that utilize 10\% to 20\% of the training labels.

Beyond overall pixel-wise accuracy (Dice), a deeper analysis of the metrics reveals our method's fundamental advantage in preserving structural integrity. Specifically, our method achieves the lowest 95\% Hausdorff Distance (HD95) across all eight datasets by a substantial margin. For instance, on challenging datasets like ORVS and DRHAGIS, ESRVS reduces HD95 relative to the strongest baseline from 125.94 to 35.06 and from 217.52 to 102.07, corresponding to reductions of 72.2\% and 53.1\%, respectively. This overwhelming superiority in HD95 indicates that our approach significantly suppresses severe outliers and false positive predictions in background regions. Furthermore, the topological continuity (clDice) results provide crucial insights into the quality of the segmented vascular networks. On the STARE dataset, although our Dice score is marginally lower than KnowSAM (0.6357 vs. 0.6547), our method yields a dramatically higher clDice (0.7283 vs. 0.6351) alongside a much tighter boundary (HD95: 19.58 vs. 29.54). Physically, this implies that while semi-supervised methods might capture thicker, disconnected vascular blobs (inflating the Dice score), our single-label method successfully recovers finer, continuous topological structures without over-segmentation. This phenomenon is consistently observed across the datasets, indicating that our framework not only minimizes annotation dependency but also produces geometrically more precise and structurally faithful vessel segmentations.

\begin{table}[t]
\centering
\caption{Segmentation performance of Mask2Former, UNet, and HSNet under two supervision settings: fully supervised training (all images labeled) and our single-label setting (ESRVS; a single annotated image). For Dice and clDice, values in parentheses denote the percentage of the fully supervised score retained by ESRVS. For HD95, values in parentheses report the absolute difference from the fully supervised model, where positive values indicate higher boundary error.
}
\label{tab:full_vs_oneshot_models}
\resizebox{\textwidth}{!}{%
\begin{tabular}{cc||cc|cc|cc}
\toprule
&& \multicolumn{2}{c|}{\textbf{Mask2Former}} 
& \multicolumn{2}{c|}{\textbf{UNet}} 
& \multicolumn{2}{c}{\textbf{HSNet}} \\
\cmidrule(lr){3-4} \cmidrule(lr){5-6} \cmidrule(lr){7-8}
{\textbf{Dataset}} & {\textbf{Metric}} 
& \textbf{Fully-labeled} & \textbf{ESRVS} 
& \textbf{Fully-labeled} & \textbf{ESRVS} 
& \textbf{Fully-labeled} & \textbf{ESRVS} \\
\midrule

\multirow{3}{*}{DRIVE} 
& \textbf{Dice $\uparrow$}   
        & 0.8252 & 0.8014 (97.12\%)
        & 0.7971 & 0.7102 (89.10\%) 
        & 0.7821 & 0.7355 (94.05\%) \\
& \textbf{clDice $\uparrow$} 
        & 0.8065 & 0.7797 (96.68\%)
        & 0.8312 & 0.6881 (82.78\%) 
        & 0.8423 & 0.8181 (97.13\%) \\
& \textbf{HD95 $\downarrow$}   
        & 6.44 & 9.59 (+3.15)
        & 4.22 & 9.55 (+5.33) 
        & 3.63 & 6.15 (+2.52)   \\
\midrule

\multirow{3}{*}{FIVES} 
        & \textbf{Dice $\uparrow$}   
        & 0.8793 & 0.7044(80.11\%)
        & 0.8749 & 0.6668(76.21\%) 
        & 0.8851 & 0.6131(69.27\%) 
          \\
& \textbf{clDice $\uparrow$} 
        & 0.9027 & 0.7735(85.69\%)
        & 0.8881 & 0.6771(76.24\%) 
        & 0.9102 & 0.7338(80.62\%) \\
& \textbf{HD95 $\downarrow$}
        & 32.13 & 76.99(+44.86)  
        & 40.81 & 64.56 (+23.75) 
        & 29.88 & 120.10(+90.22) \\
\midrule

\multirow{3}{*}{ORVS} 
& \textbf{Dice $\uparrow$}
        & 0.7213 & 0.6908(95.77\%) 
        & 0.7083 & 0.5415(76.45\%) 
        & 0.7005 & 0.5233(74.70\%) \\
& \textbf{clDice $\uparrow$}
        & 0.7115 & 0.6923(97.30\%)
        & 0.7435 & 0.6626(89.12\%) 
        & 0.7767 & 0.7374(94.94\%) \\
& \textbf{HD95 $\downarrow$}
        & 30.71 & 53.66(+22.95)  
        & 17.53 & 35.06(+17.53) 
        & 18.09 & 30.52(+12.43) \\
\midrule

\multirow{3}{*}{STARE} 
& \textbf{Dice $\uparrow$} 
        & 0.8423 & 0.8078(95.90\%) 
        & 0.7809 & 0.6357(81.40\%) 
        & 0.7511 & 0.6397(85.17\%) \\
& \textbf{clDice $\uparrow$} 
        & 0.8571 & 0.8348(97.40\%) 
        & 0.8606 & 0.7283(84.63\%)
        & 0.8766 & 0.8281(94.47\%)\\
& \textbf{HD95 $\downarrow$}
        & 4.14 & 6.56(+2.42)  
        & 5.93 & 19.58(+13.65)
        & 4.08 & 11.08(+7.00) \\
\midrule

\multirow{3}{*}{HRF} 
& \textbf{Dice $\uparrow$}
        & 0.7523 & 0.7263(96.54\%)
        & 0.7687 & 0.6347(82.57\%) 
        & 0.7342 & 0.5652(76.98\%)\\
& \textbf{clDice $\uparrow$} 
        & 0.7005 & 0.6565(93.72\%) 
        & 0.7952 & 0.6444(81.04\%)
        & 0.7924 & 0.7256(91.57\%)\\
& \textbf{HD95 $\downarrow$}
        & 59.49 & 93.29(+33.80)
        & 23.63 & 44.34(+20.71)
        & 26.04 & 61.97(+35.93)\\
\midrule

\multirow{3}{*}{CHASEDB1} 
& \textbf{Dice $\uparrow$} 
        & 0.8165 & 0.7891 (96.64\%) 
        & 0.7934 & 0.6904 (87.02\%) 
        & 0.7976 & 0.7037 (88.22\%) \\
& \textbf{clDice $\uparrow$}
        & 0.7969 & 0.7841 (98.39\%) 
        & 0.8128 & 0.6553 (80.62\%) 
        & 0.8412 & 0.8072 (95.96\%) \\
& \textbf{HD95 $\downarrow$} 
        & 17.71 & 20.18 (+2.47) 
        & 11.30 & 14.78 (+3.48) 
        & 8.84 & 15.98 (+7.14)\\
\midrule

\multirow{3}{*}{DRHAGIS} 
& \textbf{Dice $\uparrow$} 
        & 0.7329 & 0.6852(93.49\%) 
        & 0.7214 & 0.5841(80.97\%) 
        & 0.6818 & 0.4811(70.56\%)\\
& \textbf{clDice $\uparrow$} 
        & 0.7091 & 0.6826(96.26\%) 
        & 0.7736 & 0.6256(80.87\%)
        & 0.7842 & 0.6536(83.35\%)\\
& \textbf{HD95 $\downarrow$}  
        & 62.36 & 93.06(+30.70)  
        & 37.27 & 102.07(+64.8)
        & 36.36 & 87.11(+50.75)\\
\midrule

\multirow{3}{*}{AVRDB} 
& \textbf{Dice $\uparrow$}
        & 0.7171 & 0.6752 (94.16\%)
        & 0.7391 & 0.6387 (86.42\%) 
        & 0.6983 & 0.6376 (91.31\%) \\
& \textbf{clDice $\uparrow$}
        & 0.7829 & 0.7492 (95.70\%)
        & 0.6818 & 0.6891 (101.07\%) 
        & 0.7987 & 0.7349 (92.01\%) \\
& \textbf{HD95 $\downarrow$}   
        & 22.83 & 30.15 (+7.32)
        & 24.79 & 24.18 (-0.61)
        & 22.40 & 30.19 (+7.79) \\
\bottomrule
\end{tabular}
}
\end{table}

\subsection{Approaching the Fully Supervised Performance} 
The robustness of our single-label method (ESRVS) is remarkably demonstrated by comparing it against the fully supervised baseline across multiple backbone architectures. As shown in Table~\ref{tab:full_vs_oneshot_models}, ESRVS is highly compatible with the three evaluated backbones, functioning effectively on both Transformer (Mask2Former) and CNN (UNet, HSNet) paradigms. Most notably, when applied to Mask2Former, the performance gap between a single labeled image and the entire annotated dataset is substantially narrowed on most datasets. ESRVS retains nearly 95\% of the fully supervised Dice and clDice performance on the majority of datasets. For instance, on CHASEDB1, our method recovers 98.39\% of the clDice with only a marginal (+2.47) increase in HD95. For UNet and HSNet backbones, ESRVS consistently maintains a substantial retention rate, typically ranging from 80\% to 95\%. The consistently high clDice retention across all backbones highlights a fundamental strength of ESRVS: even under extreme label scarcity, it prioritizes the extraction of continuous vascular topology over mere pixel matching, rendering it a highly promising alternative to expensive fully-supervised training in clinical settings.

\begin{figure*}[t]
    \centering
    \includegraphics[width=0.90\textwidth]{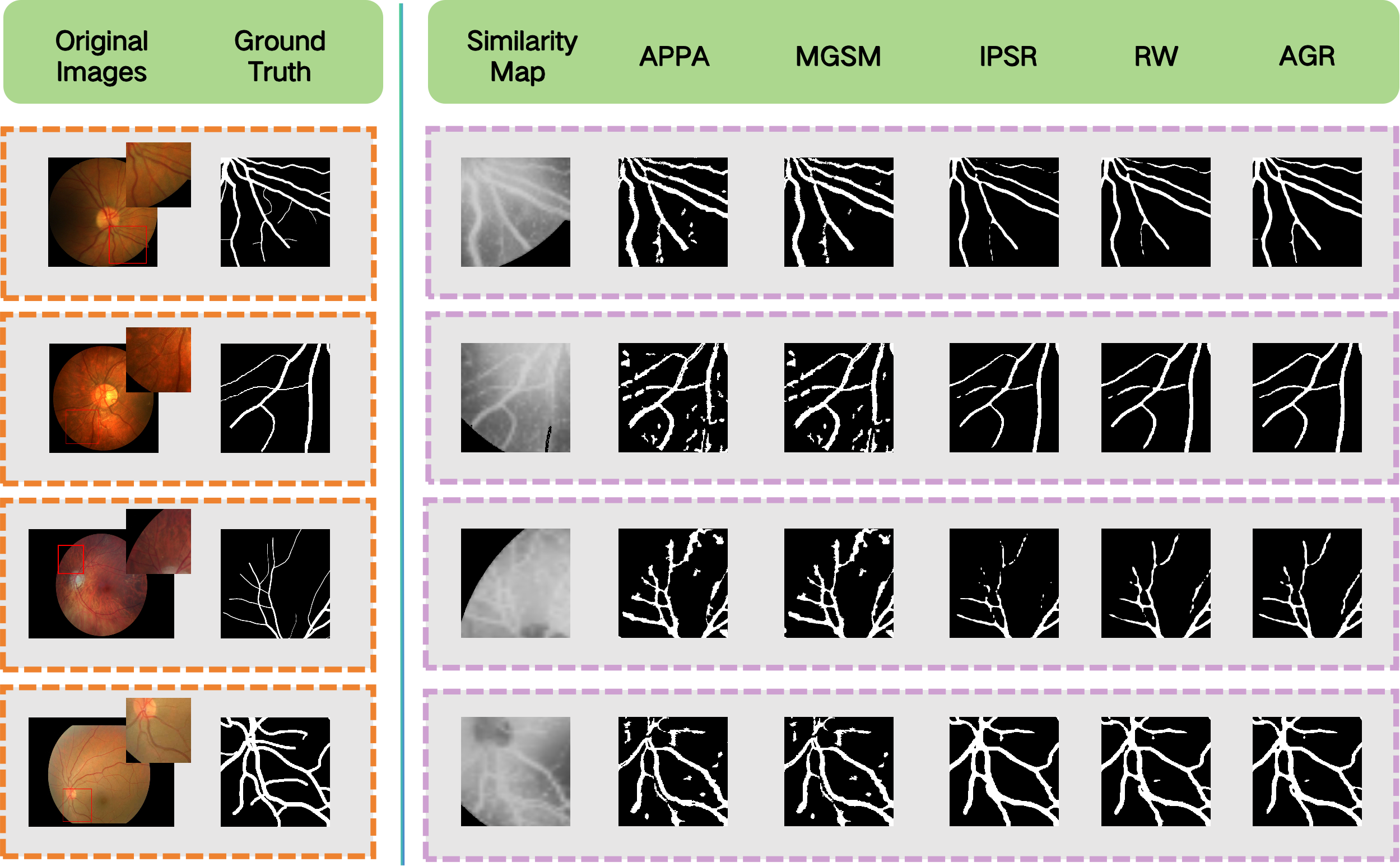}
    \caption{Qualitative ablation results with progressively added components (left to right). Rows (top to bottom) correspond to the datasets ORVS, DRHAGIS, CHASEDB1, and AVRDB.}
    \label{fig:visualization}
\end{figure*}

\begin{table}[t]
\centering
\caption{Ablation study demonstrating the contribution of each model component. Abbreviations: SSP (Self-Supervised Pretrain), IPSR (Initial Pseudo-Supervised Refinement), RW (Ref. Weighting), AGR (Adversarial-Guided Refinement).}
\label{tab:ablation}
\resizebox{\textwidth}{!}{%
\setlength{\tabcolsep}{4pt}
\begin{tabular}{@{}c cccccc ccc ccc ccc ccc@{}}
\toprule
\multirow{2}{*}{\textbf{\#}} & \multicolumn{6}{c}{\textbf{Components}} & \multicolumn{3}{c}{\textbf{DRIVE}} & \multicolumn{3}{c}{\textbf{FIVES}} & \multicolumn{3}{c}{\textbf{ORVS}} & \multicolumn{3}{c}{\textbf{STARE}} \\
\cmidrule(lr){2-7} \cmidrule(lr){8-10} \cmidrule(lr){11-13} \cmidrule(lr){14-16} \cmidrule(lr){17-19}
& SSP & APPA & MGSM & IPSR & RW & AGR 
& Dice $\uparrow$ & clDice $\uparrow$ & HD95 $\downarrow$ 
& Dice $\uparrow$ & clDice $\uparrow$ & HD95 $\downarrow$ 
& Dice $\uparrow$ & clDice $\uparrow$ & HD95 $\downarrow$ 
& Dice $\uparrow$ & clDice $\uparrow$ & HD95 $\downarrow$ \\
\midrule
1 & $\times$ & $\times$ & $\times$ & $\times$ & $\times$ & $\times$ 
& 0.4062 & 0.4792 & 50.43 
& 0.3362 & 0.3906 & 283.68 
& 0.2747 & 0.3799 & 190.31 
& 0.4638 & 0.5417 & 54.66 \\

2 & $\checkmark$ & $\times$ & $\times$ & $\times$ & $\times$ & $\times$ 
& 0.4036 & 0.4822 & 49.37 
& 0.3186 & 0.3931 & 247.41 
& 0.2831 & 0.3834 & 195.73 
& 0.4694 & 0.5441 & 54.49 \\

3 & $\checkmark$ & $\checkmark$ & $\times$ & $\times$ & $\times$ & $\times$ 
& 0.6894 & 0.7203 & 12.20 
& 0.5618 & 0.5688 & 149.30 
& 0.4567 & 0.5259 & 77.70 
& 0.6007 & 0.6903 & 23.81 \\

4 & $\checkmark$ & $\checkmark$ & $\checkmark$ & $\times$ & $\times$ & $\times$ 
& 0.6889 & 0.7236 & 12.27 
& 0.5858 & 0.5891 & 147.37 
& 0.4783 & 0.5397 & 75.50 
& 0.6017 & 0.6996 & 23.94 \\

5 & $\checkmark$ & $\checkmark$ & $\checkmark$ & $\checkmark$ & $\times$ & $\times$ 
& 0.7585 & \textbf{\underline{0.8019}} & \textbf{\underline{7.10}} 
& 0.6907 & 0.7671 & 78.81 
& 0.6331 & \textbf{\underline{0.7165}} & 45.72 
& 0.7726 & 0.8217 & 7.88 \\

6 & $\checkmark$ & $\checkmark$ & $\checkmark$ & $\checkmark$ & $\checkmark$ & $\times$ 
& 0.7838 & 0.7741 & 12.17 
& 0.6559 & 0.6715 & 111.70 
& 0.6699 & 0.7156 & \textbf{\underline{43.17}} 
& 0.7889 & 0.7881 & 9.68 \\

7 & $\checkmark$ & $\checkmark$ & $\checkmark$ & $\checkmark$ & $\checkmark$ & $\checkmark$ 
& \textbf{\underline{0.8014}} & 0.7797 & 9.59 
& \textbf{\underline{0.7044}} & \textbf{\underline{0.7735}} & \textbf{\underline{76.99}} 
& \textbf{\underline{0.6908}} & 0.6923 & 53.66 
& \textbf{\underline{0.8078}} & \textbf{\underline{0.8348}} & \textbf{\underline{6.56}} \\

\midrule
\midrule
\multirow{2}{*}{\textbf{\#}} & \multicolumn{6}{c}{\textbf{Components}} & \multicolumn{3}{c}{\textbf{HRF}} & \multicolumn{3}{c}{\textbf{CHASEDB1}} & \multicolumn{3}{c}{\textbf{DRHAGIS}} & \multicolumn{3}{c}{\textbf{AVRDB}} \\
\cmidrule(lr){2-7} \cmidrule(lr){8-10} \cmidrule(lr){11-13} \cmidrule(lr){14-16} \cmidrule(lr){17-19}
& SSP & APPA & MGSM & IPSR & RW & AGR 
& Dice $\uparrow$ & clDice $\uparrow$ & HD95 $\downarrow$ 
& Dice $\uparrow$ & clDice $\uparrow$ & HD95 $\downarrow$ 
& Dice $\uparrow$ & clDice $\uparrow$ & HD95 $\downarrow$ 
& Dice $\uparrow$ & clDice $\uparrow$ & HD95 $\downarrow$ \\
\midrule

1 & $\times$ & $\times$ & $\times$ & $\times$ & $\times$ & $\times$ 
& 0.2573 & 0.3411 & 387.21 
& 0.2967 & 0.3648 & 122.71 
& 0.2453 & 0.3426 & 228.97 
& 0.3324 & 0.4249 & 110.30 \\

2 & $\checkmark$ & $\times$ & $\times$ & $\times$ & $\times$ & $\times$ 
& 0.2576 & 0.3416 & 388.05 
& 0.2981 & 0.3685 & 119.74 
& 0.2391 & 0.3386 & 215.61 
& 0.3381 & 0.4269 & 119.64 \\

3 & $\checkmark$ & $\checkmark$ & $\times$ & $\times$ & $\times$ & $\times$ 
& 0.5363 & 0.6053 & 61.95 
& 0.6055 & 0.6425 & 32.73 
& 0.4691 & 0.5434 & 104.60 
& 0.5929 & 0.6256 & 38.30 \\

4 & $\checkmark$ & $\checkmark$ & $\checkmark$ & $\times$ & $\times$ & $\times$ 
& 0.5312 & 0.6205 & 61.51 
& 0.6228 & 0.6553 & 32.55 
& 0.4738 & 0.5614 & 105.77 
& 0.5961 & 0.6273 & 39.25 \\

5 & $\checkmark$ & $\checkmark$ & $\checkmark$ & $\checkmark$ & $\times$ & $\times$ 
& 0.6434 & \textbf{\underline{0.7339}} & \textbf{\underline{57.65}} 
& 0.7472 & \textbf{\underline{0.8007}} & \textbf{\underline{14.39}} 
& 0.6567 & 0.6353 & 100.94 
& 0.6447 & 0.7283 & \textbf{\underline{26.97}} \\

6 & $\checkmark$ & $\checkmark$ & $\checkmark$ & $\checkmark$ & $\checkmark$ & $\times$ 
& 0.7251 & 0.6534 & 98.64 
& \textbf{\underline{0.7902}} & 0.7711 & 21.65 
& 0.6561 & 0.6819 & \textbf{\underline{88.39}} 
& 0.6689 & 0.7467 & 30.13 \\

7 & $\checkmark$ & $\checkmark$ & $\checkmark$ & $\checkmark$ & $\checkmark$ & $\checkmark$ 
& \textbf{\underline{0.7263}} & 0.6565 & 93.29 
& 0.7891 & 0.7841 & 20.18 
& \textbf{\underline{0.6852}} & \textbf{\underline{0.6826}} & 93.06 
& \textbf{\underline{0.6752}} & \textbf{\underline{0.7492}} & 30.15 \\
\bottomrule
\end{tabular}
}
\end{table}

\begin{table}[t]
\centering
\caption{Cross-dataset performance comparison on the STARE dataset. Each row marked ``Ours'' uses one reference image and its corresponding label from the specified Reference Dataset to propagate annotations to the Training Dataset.}
\label{tab:Cross-dataset}
\resizebox{0.75\textwidth}{!}{%
\begin{tabular}{l l l c c c}
\hline
\textbf{Training Setting} & \textbf{Reference} & \textbf{Training Dataset} & \textbf{Dice} & \textbf{clDice} & \textbf{HD95} \\
\hline
Fully-Supervised & —— & STARE (all labeled) & \textbf{\underline{0.8423}} & \textbf{\underline{0.8571}} & \textbf{\underline{4.14}} \\
\hline
\multirow{8}{*}{\makecell{Single Label \\ (Ours)}}
 & STARE    & STARE (unlabeled) & \textbf{\underline{0.8078}} & 0.8348 & 6.56 \\
 & CHASEDB1 & STARE  (unlabeled) & 0.7905 & 0.8146 & 9.41 \\
 & ORVS     & STARE  (unlabeled) & 0.7851 & 0.7867 & 10.38 \\
 & FIVES    & STARE  (unlabeled) & 0.8023 & 0.8146 & 7.99 \\
 & HRF      & STARE  (unlabeled) & 0.7867 & 0.8077 & 9.36 \\
 & AVRDB    & STARE  (unlabeled) & 0.7749 & \textbf{\underline{0.8446}} & \textbf{\underline{6.31}} \\
 & DRHAGIS  & STARE  (unlabeled) & 0.7907 & 0.8045 & 11.91 \\
 & DRIVE    & STARE  (unlabeled) & 0.7944 & 0.8148 & 8.66 \\
\hline
\end{tabular}
}
\end{table}

\subsection{Ablation Studies and Component Contributions} 
All ablation and cross-dataset experiments are conducted using Mask2Former as the segmentation backbone. Table~\ref{tab:ablation} shows that APPA and IPSR provide the largest improvements. Notably, the introduction of the APPA module (Setup 2 to 3) triggers a profound leap in baseline performance, demonstrating the importance of the green-channel structural prior for initial pseudo-label generation. For instance, on the STARE dataset, clDice surges from 0.5441 to 0.6903, and HD95 drops by more than half. MGSM provides smaller but generally positive improvements in Dice and clDice by enriching the reference representation across scales. IPSR further produces large gains, particularly in HD95, indicating that the first pseudo-supervised stage removes many distant false positives. Reference Weighting and Adversarial-Guided Refinement improve Dice on most datasets, although clDice and HD95 exhibit dataset-dependent trade-offs. Overall, the components make complementary contributions rather than producing monotonic improvement on every metric, as shown in the ablation visualization in Fig.~\ref{fig:visualization}. 

One of the most formidable challenges in medical image analysis is domain shift caused by varying scanner protocols and patient demographics. To evaluate the robustness of ESRVS against such shifts, we conduct a cross-dataset evaluation where the network segments unlabeled STARE images using a single reference image sampled from completely different datasets. As presented in Table~\ref{tab:Cross-dataset}, our method exhibits exceptional zero-shot cross-domain adaptability. Astonishingly, relying on a reference image from a different acquisition distribution (e.g., AVRDB or FIVES) barely degrades performance compared to using an in-domain STARE reference. Most impressively, when using one AVRDB image as the reference, the model achieves a clDice of 0.8446 and an HD95 of 6.31 on STARE—scores that virtually rival the fully-supervised performance (clDice: 0.8571, HD95: 4.14). This compelling evidence demonstrates that ESRVS successfully learns transferable representations of vascular geometry, rather than overfitting to the domain-specific intensity or color distributions of the reference image.

\section{Limitations}
ESRVS relies on access to an unlabeled target-domain image pool and on selecting a reference image that is representative in feature space. Its green-channel prior may be less reliable under severe illumination changes, atypical acquisition modalities, or pathologies that substantially alter vessel appearance. Although the stage-wise refinement process reduces pseudo-label noise, systematic errors may still be reinforced during training. Finally, our experiments evaluate segmentation performance on public benchmarks and do not establish downstream clinical utility.

\section{Conclusion}
We presented ESRVS, a framework that advances label-efficient segmentation using only one annotated image. By integrating foundation model-driven propagation, physics-aware priors, and progressive denoising, ESRVS substantially mitigates the fragmentation and topological errors typical of extreme low-data regimes. Compared with the semi-supervised baselines, ESRVS achieves the best Dice and clDice on six of eight datasets, and the best HD95 on all eight datasets, despite using only one annotated image. Furthermore, its remarkable cross-dataset generalization highlights its robustness against domain shifts. Our results suggest that high-quality, continuous vascular segmentation can be achieved without relying on large amounts of expensive dense annotations, offering a promising solution for deploying medical imaging AI in resource-constrained clinical environments.

\textbf{Acknowledgment.} This research was supported in part by the Natural Science Foundation of Jiangsu Province (Grant BK20220949) and National Natural Science Foundation of China (Grant 62201263).

\bibliographystyle{splncs04}
\bibliography{reference}

@article{qin2024review,
  title={A review of retinal vessel segmentation for fundus image analysis},
  author={Qin, Qing and Chen, Yuanyuan},
  journal={Engineering Applications of Artificial Intelligence},
  volume={128},
  pages={107454},
  year={2024},
  publisher={Elsevier}
}

@article{kellner2024eye,
  title={The eye as the window to the heart: optical coherence tomography angiography biomarkers as indicators of cardiovascular disease},
  author={Kellner, Rebecca L and Harris, Alon and Ciulla, Lauren and Guidoboni, Giovanna and Verticchio Vercellin, Alice and Oddone, Francesco and Carnevale, Carmela and Zaid, Mohamed and Antman, Gal and Kuvin, Jeffrey T and others},
  journal={Journal of Clinical Medicine},
  volume={13},
  number={3},
  pages={829},
  year={2024},
  publisher={MDPI}
}

@article{yao2024cnn,
  title={From cnn to transformer: A review of medical image segmentation models},
  author={Yao, Wenjian and Bai, Jiajun and Liao, Wei and Chen, Yuheng and Liu, Mengjuan and Xie, Yao},
  journal={Journal of Imaging Informatics in Medicine},
  volume={37},
  number={4},
  pages={1529--1547},
  year={2024},
  publisher={Springer}
}

@inproceedings{zhou2018unet++,
  title={Unet++: A nested u-net architecture for medical image segmentation},
  author={Zhou, Zongwei and Rahman Siddiquee, Md Mahfuzur and Tajbakhsh, Nima and Liang, Jianming},
  booktitle={International workshop on deep learning in medical image analysis},
  pages={3--11},
  year={2018},
  organization={Springer}
}

@article{gu2019net,
  title={Ce-net: Context encoder network for 2d medical image segmentation},
  author={Gu, Zaiwang and Cheng, Jun and Fu, Huazhu and Zhou, Kang and Hao, Huaying and Zhao, Yitian and Zhang, Tianyang and Gao, Shenghua and Liu, Jiang},
  journal={IEEE transactions on medical imaging},
  volume={38},
  number={10},
  pages={2281--2292},
  year={2019},
  publisher={IEEE}
}

@article{lu2025semi,
  title={Semi-Supervised Retinal Vessel Segmentation Based on Pseudo Label Filtering},
  author={Lu, Zheng and Li, Jiaguang and Liu, Zhenyu and Cao, Qian and Tian, Tao and Wang, Xianchao and Huang, Zanjie},
  journal={Symmetry},
  volume={17},
  number={9},
  pages={1462},
  year={2025},
  publisher={MDPI}
}

@article{alsaleh2024few,
  title={Few-shot learning for medical image segmentation using 3D U-Net and model-agnostic meta-learning (MAML)},
  author={Alsaleh, Aqilah M and Albalawi, Eid and Algosaibi, Abdulelah and Albakheet, Salman S and Khan, Surbhi Bhatia},
  journal={Diagnostics},
  volume={14},
  number={12},
  pages={1213},
  year={2024},
  publisher={MDPI}
}

@article{dosovitskiy2020image,
  title={An image is worth 16x16 words: Transformers for image recognition at scale},
  author={Dosovitskiy, Alexey},
  journal={arXiv preprint arXiv:2010.11929},
  year={2020}
}

@inproceedings{caron2021emerging,
  title={Emerging properties in self-supervised vision transformers},
  author={Caron, Mathilde and Touvron, Hugo and Misra, Ishan and J{\'e}gou, Herv{\'e} and Mairal, Julien and Bojanowski, Piotr and Joulin, Armand},
  booktitle={Proceedings of the IEEE/CVF international conference on computer vision},
  pages={9650--9660},
  year={2021}
}

@article{oquab2023dinov2,
  title={Dinov2: Learning robust visual features without supervision},
  author={Oquab, Maxime and Darcet, Timoth{\'e}e and Moutakanni, Th{\'e}o and Vo, Huy and Szafraniec, Marc and Khalidov, Vasil and Fernandez, Pierre and Haziza, Daniel and Massa, Francisco and El-Nouby, Alaaeldin and others},
  journal={arXiv preprint arXiv:2304.07193},
  year={2023}
}

@article{simeoni2025dinov3,
  title={Dinov3},
  author={Sim{\'e}oni, Oriane and Vo, Huy V and Seitzer, Maximilian and Baldassarre, Federico and Oquab, Maxime and Jose, Cijo and Khalidov, Vasil and Szafraniec, Marc and Yi, Seungeun and Ramamonjisoa, Micha{\"e}l and others},
  journal={arXiv preprint arXiv:2508.10104},
  year={2025}
}

@article{wu2021scs,
  title={Scs-net: A scale and context sensitive network for retinal vessel segmentation},
  author={Wu, Huisi and Wang, Wei and Zhong, Jiafu and Lei, Baiying and Wen, Zhenkun and Qin, Jing},
  journal={Medical Image Analysis},
  volume={70},
  pages={102025},
  year={2021},
  publisher={Elsevier}
}

@inproceedings{zhang2025towards,
  title={Towards Robust Retinal Vessel Segmentation via Reducing Open-Set Label Noises from SAM-Generated Masks},
  author={{Zhang, Minqing} and He, Mengxian and Yuan, Wu},
  booktitle={International Conference on Medical Image Computing and Computer-Assisted Intervention},
  pages={631--641},
  year={2025},
  organization={Springer}
}

@inproceedings{ronneberger2015u,
  title={U-net: Convolutional networks for biomedical image segmentation},
  author={Ronneberger, Olaf and Fischer, Philipp and Brox, Thomas},
  booktitle={International Conference on Medical image computing and computer-assisted intervention},
  pages={234--241},
  year={2015},
  organization={Springer}
}

@article{liu2023wave,
  title={Wave-Net: A lightweight deep network for retinal vessel segmentation from fundus images},
  author={Liu, Yanhong and Shen, Ji and Yang, Lei and Yu, Hongnian and Bian, Guibin},
  journal={Computers in biology and medicine},
  volume={152},
  pages={106341},
  year={2023},
  publisher={Elsevier}
}

@article{jiang2024covi,
  title={CoVi-Net: A hybrid convolutional and vision transformer neural network for retinal vessel segmentation},
  author={Jiang, Minshan and Zhu, Yongfei and Zhang, Xuedian},
  journal={Computers in Biology and Medicine},
  volume={170},
  pages={108047},
  year={2024},
  publisher={Elsevier}
}

@article{tan2023oct2former,
  title={OCT2Former: A retinal OCT-angiography vessel segmentation transformer},
  author={Tan, Xiao and Chen, Xinjian and Meng, Qingquan and Shi, Fei and Xiang, Dehui and Chen, Zhongyue and Pan, Lingjiao and Zhu, Weifang},
  journal={Computer Methods and Programs in Biomedicine},
  volume={233},
  pages={107454},
  year={2023},
  publisher={Elsevier}
}

@article{moezzi2023uncertainty,
  title={An uncertainty-aware pseudo-label selection framework using regularized conformal prediction},
  author={Moezzi, Matin},
  journal={arXiv preprint arXiv:2309.15963},
  year={2023}
}

@article{zheng2025crossnext,
  title={CrossNeXt: ConvNeXt-based cross-teaching with entropy minimization for semi-supervised liver segmentation from abdominal MRI},
  author={Zheng, Zhiji and Luo, Xiao and Li, Peiwen and Piao, Sirong and Cao, Xin and Liu, Xiao and Yang, Liqin and Hu, Bin and Geng, Yan and Geng, Daoying},
  journal={Computerized Medical Imaging and Graphics},
  pages={102624},
  year={2025},
  publisher={Elsevier}
}

@inproceedings{ouali2020semi,
  title={Semi-supervised semantic segmentation with cross-consistency training},
  author={Ouali, Yassine and Hudelot, C{\'e}line and Tami, Myriam},
  booktitle={Proceedings of the IEEE/CVF conference on computer vision and pattern recognition},
  pages={12674--12684},
  year={2020}
}

@article{you2022fmwdct,
  title={FMWDCT: Foreground mixup into weighted dual-network cross training for semisupervised remote sensing road extraction},
  author={{You, Zhi-Hui} and Wang, Jia-Xin and Chen, Si-Bao and Tang, Jin and Luo, Bin},
  journal={IEEE Journal of Selected Topics in Applied Earth Observations and Remote Sensing},
  volume={15},
  pages={5570--5579},
  year={2022},
  publisher={IEEE}
}

@article{zhao2024semi,
  title={Semi-supervised medical image segmentation based on deep consistent collaborative learning},
  author={Zhao, Xin and Wang, Wenqi},
  journal={Journal of Imaging},
  volume={10},
  number={5},
  pages={118},
  year={2024},
  publisher={MDPI}
}

@article{staal2004ridge,
  title={Ridge-based vessel segmentation in color images of the retina},
  author={Staal, Joes and Abr{\`a}moff, Michael D and Niemeijer, Meindert and Viergever, Max A and Van Ginneken, Bram},
  journal={IEEE transactions on medical imaging},
  volume={23},
  number={4},
  pages={501--509},
  year={2004},
  publisher={IEEE}
}

@article{hoover2000locating,
  title={Locating blood vessels in retinal images by piecewise threshold probing of a matched filter response},
  author={Hoover, AD and Kouznetsova, Valentina and Goldbaum, Michael},
  journal={IEEE Transactions on Medical imaging},
  volume={19},
  number={3},
  pages={203--210},
  year={2000},
  publisher={IEEE}
}

@article{owen2011retinal,
  title={Retinal arteriolar tortuosity and cardiovascular risk factors in a multi-ethnic population study of 10-year-old children; the Child Heart and Health Study in England (CHASE)},
  author={Owen, Christopher G and Rudnicka, Alicja R and Nightingale, Claire M and Mullen, Robert and Barman, Sarah A and Sattar, Naveed and Cook, Derek G and Whincup, Peter H},
  journal={Arteriosclerosis, thrombosis, and vascular biology},
  volume={31},
  number={8},
  pages={1933--1938},
  year={2011},
  publisher={Lippincott Williams \& Wilkins Hagerstown, MD}
}

@article{budai2013robust,
  title={Robust vessel segmentation in fundus images},
  author={Budai, Attila and Bock, R{\"u}diger and Maier, Andreas and Hornegger, Joachim and Michelson, Georg},
  journal={International journal of biomedical imaging},
  volume={2013},
  number={1},
  pages={154860},
  year={2013},
  publisher={Wiley Online Library}
}

@article{jin2022fives,
  title={Fives: A fundus image dataset for artificial intelligence based vessel segmentation},
  author={Jin, Kai and Huang, Xingru and Zhou, Jingxing and Li, Yunxiang and Yan, Yan and Sun, Yibao and Zhang, Qianni and Wang, Yaqi and Ye, Juan},
  journal={Scientific data},
  volume={9},
  number={1},
  pages={475},
  year={2022},
  publisher={Nature Publishing Group UK London}
}

@inproceedings{sarhan2021transfer,
  title={Transfer learning through weighted loss function and group normalization for vessel segmentation from retinal images},
  author={Sarhan, Abdullah and Rokne, Jon and Alhajj, Reda and Crichton, Andrew},
  booktitle={2020 25th International Conference on Pattern Recognition (ICPR)},
  pages={9211--9218},
  year={2021},
  organization={IEEE}
}

@article{holm2017dr,
  title={DR HAGIS—a fundus image database for the automatic extraction of retinal surface vessels from diabetic patients},
  author={Holm, Sven and Russell, Greg and Nourrit, Vincent and McLoughlin, Niall},
  journal={Journal of Medical Imaging},
  volume={4},
  number={1},
  pages={014503--014503},
  year={2017},
  publisher={Society of Photo-Optical Instrumentation Engineers}
}

@inproceedings{akbar2017avrdb,
  title={AVRDB: annotated dataset for vessel segmentation and calculation of arteriovenous ratio},
  author={Akbar, Shahzad and Hassan, Taimur and Akram, M Usman and Yasin, Ubaid Ullah and Basit, Imran},
  booktitle={Proceedings of the International Conference on Image Processing, Computer Vision, and Pattern Recognition (IPCV)},
  pages={129--134},
  year={2017},
  organization={The Steering Committee of The World Congress in Computer Science, Computer~…}
}

@article{jing2020self,
  title={Self-supervised visual feature learning with deep neural networks: A survey},
  author={Jing, Longlong and Tian, Yingli},
  journal={IEEE transactions on pattern analysis and machine intelligence},
  volume={43},
  number={11},
  pages={4037--4058},
  year={2020},
  publisher={IEEE}
}

@inproceedings{wang2019panet,
  title={Panet: Few-shot image semantic segmentation with prototype alignment},
  author={Wang, Kaixin and Liew, Jun Hao and Zou, Yingtian and Zhou, Daquan and Feng, Jiashi},
  booktitle={proceedings of the IEEE/CVF international conference on computer vision},
  pages={9197--9206},
  year={2019}
}

@inproceedings{ouyang2020self,
  title={Self-supervision with superpixels: Training few-shot medical image segmentation without annotation},
  author={Ouyang, Cheng and Biffi, Carlo and Chen, Chen and Kart, Turkay and Qiu, Huaqi and Rueckert, Daniel},
  booktitle={European conference on computer vision},
  pages={762--780},
  year={2020},
  organization={Springer}
}

@article{sun2026multi,
  title={Multi-level hierarchical prototype for few-shot medical image segmentation},
  author={Sun, Song and Yang, Jinzhu and Tang, Lingzhi and Feng, Yong and Yu, Qi},
  journal={Biomedical Signal Processing and Control},
  volume={112},
  pages={108500},
  year={2026},
  publisher={Elsevier}
}

@article{tarvainen2017mean,
  title={Mean teachers are better role models: Weight-averaged consistency targets improve semi-supervised deep learning results},
  author={Tarvainen, Antti and Valpola, Harri},
  journal={Advances in neural information processing systems},
  volume={30},
  year={2017}
}

@article{verma2022interpolation,
  title={Interpolation consistency training for semi-supervised learning},
  author={Verma, Vikas and Kawaguchi, Kenji and Lamb, Alex and Kannala, Juho and Solin, Arno and Bengio, Yoshua and Lopez-Paz, David},
  journal={Neural Networks},
  volume={145},
  pages={90--106},
  year={2022},
  publisher={Elsevier}
}

@inproceedings{chen2021semi,
  title={Semi-supervised semantic segmentation with cross pseudo supervision},
  author={Chen, Xiaokang and Yuan, Yuhui and Zeng, Gang and Wang, Jingdong},
  booktitle={Proceedings of the IEEE/CVF conference on computer vision and pattern recognition},
  pages={2613--2622},
  year={2021}
}

@inproceedings{luo2022semi,
  title={Semi-supervised medical image segmentation via cross teaching between cnn and transformer},
  author={Luo, Xiangde and Hu, Minhao and Song, Tao and Wang, Guotai and Zhang, Shaoting},
  booktitle={International conference on medical imaging with deep learning},
  pages={820--833},
  year={2022},
  organization={PMLR}
}

@inproceedings{luo2021efficient,
  title={Efficient semi-supervised gross target volume of nasopharyngeal carcinoma segmentation via uncertainty rectified pyramid consistency},
  author={Luo, Xiangde and Liao, Wenjun and Chen, Jieneng and Song, Tao and Chen, Yinan and Zhang, Shichuan and Chen, Nianyong and Wang, Guotai and Zhang, Shaoting},
  booktitle={International Conference on Medical Image Computing and Computer-Assisted Intervention},
  pages={318--329},
  year={2021},
  organization={Springer}
}

@article{huang2025learnable,
  title={Learnable prompting sam-induced knowledge distillation for semi-supervised medical image segmentation},
  author={Huang, Kaiwen and Zhou, Tao and Fu, Huazhu and Zhang, Yizhe and Zhou, Yi and Gong, Chen and Liang, Dong},
  journal={IEEE Transactions on Medical Imaging},
  year={2025},
  publisher={IEEE}
}

@article{zhang2022hsnet,
  title={HSNet: A hybrid semantic network for polyp segmentation},
  author={Zhang, Wenchao and Fu, Chong and Zheng, Yu and Zhang, Fangyuan and Zhao, Yanli and Sham, Chiu-Wing},
  journal={Computers in biology and medicine},
  volume={150},
  pages={106173},
  year={2022},
  publisher={Elsevier}
}

@inproceedings{cheng2022masked,
  title={Masked-attention mask transformer for universal image segmentation},
  author={Cheng, Bowen and Misra, Ishan and Schwing, Alexander G and Kirillov, Alexander and Girdhar, Rohit},
  booktitle={Proceedings of the IEEE/CVF conference on computer vision and pattern recognition},
  pages={1290--1299},
  year={2022}
}

@article{ricci2007retinal,
  title={Retinal blood vessel segmentation using line operators and support vector classification},
  author={Ricci, Elisa and Perfetti, Renzo},
  journal={IEEE transactions on medical imaging},
  volume={26},
  number={10},
  pages={1357--1365},
  year={2007},
  publisher={IEEE}
}

@inproceedings{zhang2026medical,
  title={Medical image segmentation with minimal labeling effort: how far can we push the limits?},
  author={Zhang, Yizhe},
  booktitle={Proceedings of the AAAI Conference on Artificial Intelligence},
  pages={28492--28500},
  year={2026}
}

@article{dice1945measures,
  title={Measures of the amount of ecologic association between species},
  author={Dice, Lee R},
  journal={Ecology},
  volume={26},
  number={3},
  pages={297--302},
  year={1945},
  publisher={JSTOR}
}

@inproceedings{shit2021cldice,
  title={clDice-a novel topology-preserving loss function for tubular structure segmentation},
  author={Shit, Suprosanna and Paetzold, Johannes C and Sekuboyina, Anjany and Ezhov, Ivan and Unger, Alexander and Zhylka, Andrey and Pluim, Josien PW and Bauer, Ulrich and Menze, Bjoern H},
  booktitle={Proceedings of the IEEE/CVF conference on computer vision and pattern recognition},
  pages={16560--16569},
  year={2021}
}

@article{taha2015metrics,
  title={Metrics for evaluating 3D medical image segmentation: analysis, selection, and tool},
  author={Taha, Abdel Aziz and Hanbury, Allan},
  journal={BMC medical imaging},
  volume={15},
  number={1},
  pages={29},
  year={2015},
  publisher={Springer}
}
\end{document}


\appendix

\section{Dataset Details}
We evaluate our method on eight public fundus vessel segmentation datasets, covering a wide range of imaging conditions, resolutions, and pathological variations. Among the commonly used benchmarks, DRIVE contains 40 fundus images with manual vessel annotations, while STARE consists of 20 images including several pathological cases. CHASEDB1 includes 28 images collected from a child health study, and HRF provides 45 high-resolution images spanning both healthy subjects and patients with diabetic retinopathy and glaucoma. In addition, FIVES is a large-scale dataset with 800 images and substantial intra-dataset diversity. We further include three specialized datasets to evaluate robustness under more challenging clinical conditions. ORVS contains 49 images and focuses on vessel segmentation under domain shift scenarios. DRHAGIS consists of 40 images for diabetic retinopathy analysis, featuring hemorrhages and complex pathological structures. AVRDB contains 100 annotated images with artery-vein labels, enabling more fine-grained vascular analysis. The specific dataset splits are summarized in Table~\ref{tab:data_splits}.

\begin{table}[h]
\footnotesize
\centering
\caption{Data splits under different training settings for each dataset. All methods are evaluated on the same test sets.}
\label{tab:data_splits}
\resizebox{0.9\textwidth}{!}{
\begin{tabular}{l cc | cc | cc | cc}
\toprule

& & & \multicolumn{2}{c}{\textbf{Fully-supervised}} 
& \multicolumn{2}{c}{\textbf{Semi-supervised}} 
& \multicolumn{2}{c}{\textbf{Ours}} \\
\cmidrule(lr){4-5} \cmidrule(lr){6-7} \cmidrule(lr){8-9}

\textbf{Dataset} 
& \textbf{Total} & \textbf{Training}
& \textbf{Labeled} & \textbf{Unlabeled}
& \textbf{Labeled} & \textbf{Unlabeled}
& \textbf{Labeled} & \textbf{Unlabeled} \\
\midrule

DRIVE & 40 & 28 & 28 & 0 & 6 & 22 & 1 & 27 \\
STARE & 20  & 14 & 14 & 0 & 3 & 11 & 1 & 13  \\
CHASEDB1 & 28 & 20& 20 & 0 & 4 & 16 & 1 & 19 \\
HRF & 45 & 32 & 32 & 0 & 6 & 26 & 1 & 31 \\
FIVES & 800& 600 & 600 & 0 & 60 & 540 & 1 & 599 \\
ORVS & 49 & 34 & 34 & 0 & 7 & 27 & 1 & 33 \\
DRHAGIS & 40 & 28 & 28 & 0 & 6 & 22 & 1 & 27 \\
AVRDB & 100 & 75& 75 & 0 & 15 & 60 & 1 & 74 \\

\bottomrule
\end{tabular}
}
\end{table}

\section{Implementation Details}
The proposed framework is implemented in PyTorch and consists of a multi-stage pipeline, including zero-shot pseudo-label generation and an iterative self-training paradigm enhanced by adversarial learning. Throughout the framework, the target image size is dynamically configured based on the stage requirements: $2048 \times 2048$ for fine-grained zero-shot prototype matching, and $1024 \times 1024$ for the segmentation network. During network optimization, the training batch size is strictly set to 1 pair per iteration (consisting of exactly one reference sample and one training sample) to maintain the one-shot constraint, while a batch size of 4 is applied during the offline pseudo-label inference phase. All experiments are conducted on a machine equipped with an NVIDIA L20 GPU and 256G RAM. We evaluate segmentation performance using Dice coefficient, clDice, and HD95, which respectively measure region overlap, topological consistency, and boundary accuracy (with lower HD95 indicating better performance).

\noindent \textbf{Initial Pseudo-Label Generation.}
All fundus images are pre-processed using physical prior enhancement to emphasize vascular structures. Specifically, each input image is reconstructed into a three-channel tensor by concatenating the inverted green channel, a Contrast Limited Adaptive Histogram Equalization (clip limit = 2.0, grid size = $8 \times 8$) enhanced green channel, and the original green channel. A pre-trained DINOv3 model is used without fine-tuning. Input images are resized to $2048 \times 2048$ for fine-grained patch token extraction. A global vessel prototype is constructed by aggregating reference features masked by the one-shot ground truth, combining both global and local ($4 \times 4$ non-overlapping tiles) feature representations. Pseudo-labels are generated by computing spatial cosine similarity between the prototype and target image features. This similarity is fused with the inverted green channel prior and amplified by an exponential factor of 1.2. An adaptive local thresholding strategy is applied using a uniform filter with a window size of $\max(15, \mathrm{target\_size}/25)$ and a margin of $+0.02$ by default.

\noindent \textbf{Pseudo-label supervised optimization.}
The self-training stage consists of up to two rounds, each trained for 10 epochs at a resolution of $1024 \times 1024$. In the first round, a Mask2Former model with a Swin-Tiny backbone is adopted. Optimization is performed using AdamW with an initial learning rate of $5 \times 10^{-5}$ and a cosine annealing schedule. A paired-batching strategy is used, where one reference sample and one training sample are processed per iteration. The core segmentation objective follows the default Mask2Former bipartite matching loss, which inherently comprises a Cross-Entropy loss for mask classification, along with Focal and Dice losses for pixel-level mask prediction. To preserve the one-shot prior, the segmentation loss of the reference sample is assigned a dominant weight of 3.0 relative to the pseudo-label training loss.

\noindent \textbf{Adversarial-guided refinement.}
In the second round, a PatchGAN-style discriminator is introduced to improve alignment between pseudo-labels and ground truth. The discriminator consists of three convolutional layers followed by Batch Normalization and LeakyReLU (negative slope = 0.2). It is trained using AdamW with learning rate $2.5 \times 10^{-5}$ and momentum parameters $\beta=(0.5, 0.999)$. Binary Cross-Entropy with Logits is used as the adversarial loss, where ground-truth masks are treated as real samples and bilinearly upsampled predictions from Mask2Former as fake samples. The adversarial loss is weighted by $\lambda_{\mathrm{adv}} = 0.10$ in the generator objective.

\noindent All architectures (e.g., DINOv3 and Mask2Former) and hyperparameters are provided as default empirical configurations used in our implementation. The detailed settings are reported in Table~\ref{tab:implementation_details}.

\begin{table}[t]
\centering
\caption{Detailed Hyperparameters and Implementation Configurations}
\label{tab:implementation_details}
\resizebox{0.9\columnwidth}{!}{
\begin{tabular}{@{}ll@{}}
\toprule
\multicolumn{2}{c}{\textit{\textbf{Phase 0: Initial Pseudo-Label Generation (DINOv3)}}} \\ \midrule
Network Architecture & Pre-trained DINOv3 \\
Input Resolution & $2048 \times 2048$ \\
Inference Batch Size & 4 \\
Local Tiling Grid & $4 \times 4$ (Non-overlapping) \\
Amplification Factor & 1.2 \\
Filter Window Size & $\max(15, \mathrm{Size}/25)$ \\
Threshold Margin & $+0.02$ \\ \midrule

\multicolumn{2}{c}{\textit{\textbf{Phase 1: Pseudo-label Supervised Optimization (Mask2Former)}}} \\ \midrule
Network Architecture & Mask2Former (Swin-Tiny backbone) \\
Input Resolution & $1024 \times 1024$ \\
Training Epochs & 10 \\
Training Batch Size & 1 Pair (1 Ref. + 1 Target) \\
Optimizer & AdamW \\
Initial Learning Rate & $5 \times 10^{-5}$ \\
Learning Rate Schedule & Cosine Annealing \\
Segmentation Loss & CE + Focal + Dice (Bipartite Matching) \\
Reference Loss Weight & 3.0 \\ \midrule

\multicolumn{2}{c}{\textit{\textbf{Phase 2: Adversarial-guided Refinement (PatchGAN)}}} \\ \midrule
Discriminator Optimizer & AdamW \\
Discriminator Learning Rate & $2.5 \times 10^{-5}$ \\
Momentum ($\beta$) & $(0.5, 0.999)$ \\
Adversarial Loss Function & BCEWithLogits \\
Adversarial Weight ($\lambda_{\mathrm{adv}}$) & 0.10 \\ 
Reference Loss Weight & 3.0 \\ \bottomrule
\end{tabular}%
}
\end{table}

\begin{table}[t]
\centering
\caption{Quantitative comparison between Centroid-based Selection and Random Selection for the one-shot reference image. Bold and underlined values indicate the best performance.}
\label{tab:ref_selection}
\resizebox{0.75\textwidth}{!}{
\begin{tabular}{@{}l|ccc|ccc@{}}
\toprule
\multirow{2}{*}{\textbf{Dataset}} & \multicolumn{3}{c|}{\textbf{Centroid-based Selection}} & \multicolumn{3}{c}{\textbf{Random Selection}} \\ \cmidrule(l){2-7} 
 & \textbf{Dice $\uparrow$} & \textbf{clDice $\uparrow$} & \textbf{HD95 $\downarrow$} & \textbf{Dice $\uparrow$} & \textbf{clDice $\uparrow$} & \textbf{HD95 $\downarrow$} \\ \midrule

\textbf{DRIVE} 
& \underline{\textbf{0.8014}} & \underline{\textbf{0.7797}} & \underline{\textbf{9.59}} 
& 0.7934 & 0.7624 & 11.81 \\

\textbf{FIVES} 
& \underline{\textbf{0.7044}} & \underline{\textbf{0.7735}} & \underline{\textbf{76.99}} 
& 0.6628 & 0.7336 & 106.38 \\

\textbf{ORVS} 
& 0.6908 & 0.6923 & 53.66 
& \underline{\textbf{0.7003}} & \underline{\textbf{0.7027}} & \underline{\textbf{45.65}} \\

\textbf{STARE} 
& \underline{\textbf{0.8078}} & \underline{\textbf{0.8348}} & \underline{\textbf{6.56}} 
& 0.7775 & 0.8103 & 11.07 \\

\textbf{HRF} 
& \underline{\textbf{0.7263}} & 0.6565 & 93.29 
& 0.7209 & \underline{\textbf{0.7066}} & \underline{\textbf{63.51}} \\

\textbf{CHASEDB1} 
& 0.7891 & \underline{\textbf{0.7841}} & \underline{\textbf{20.18}} 
& \underline{\textbf{0.7973}} & 0.7711 & 30.41 \\

\textbf{DRHAGIS} 
& \underline{\textbf{0.6852}} & \underline{\textbf{0.6826}} & \underline{\textbf{93.06}} 
& 0.6826 & 0.6234 & 120.14 \\

\textbf{AVRDB} 
& \underline{\textbf{0.6752}} & \underline{\textbf{0.7492}} & 30.15 
& 0.6688 & 0.7265 & \underline{\textbf{28.88}} \\

\bottomrule
\end{tabular}
}
\end{table}

\section{Analysis of Reference Image Selection}
To address practical clinical scenarios where users may only have access to a randomly acquired, non-ideal sample for annotation, we extensively evaluate the performance difference when a random image is selected as the one-shot reference, as opposed to our proposed centroid-based method. The results across eight datasets are detailed in Table~\ref{tab:ref_selection}.

\noindent As observed in Table~\ref{tab:ref_selection}, while the centroid-based method dominates on datasets like DRIVE, FIVES, and STARE, random selection occasionally yields better metrics on datasets such as ORVS or HRF. This phenomenon can be attributed to a "lottery effect." In datasets with relatively homogeneous image qualities, a randomly selected image might coincidentally possess specific hard-to-segment morphological features that provide stronger supervisory signals than the mathematically averaged centroid. However, such instances are purely stochastic. Despite this fluctuation, the centroid-based approach is strictly necessary for the following three critical reasons:
\begin{itemize} 
\item \textbf{Eliminating Unacceptable Variance:} Random selection is inherently stochastic. This lack of reproducibility is fatal in scientific research and clinical deployment. The centroid method provides a \textit{deterministic mathematical guarantee}, ensuring that the selected reference image is systematically identical and reproducible for any given dataset. 
\item \textbf{Prevention of Catastrophic Failure:} In medical imaging, the penalty for a worst-case scenario far outweighs the benefit of a best-case scenario. A randomly selected image has a non-zero probability of being extreme outlier (e.g., severe artifacts, extreme illumination variance, or massive pathological lesions). When an outlier is randomly selected, the one-shot prototype collapses, leading to catastrophic performance degradation. 
For instance, on the relatively large FIVES dataset, a randomly selected image is unlikely to adequately represent the overall data distribution. As a result, the performance degrades noticeably, with Dice decreasing from 0.7044 to 0.6628 and clDice from 0.7735 to 0.7336, while the HD95 error increases significantly from 76.99 to 106.38. The centroid method strictly filters out these outliers by anchoring to the core distribution. 
\item \textbf{The Unreliable "Lottery Effect":} The instances where random selection outperforms the centroid method can be attributed to stochastic luck. A randomly picked image might coincidentally contain complex vascular structures that perfectly supervise the network for that specific dataset. However, treating a medical algorithm like a lottery is unscalable. \end{itemize}

\noindent While our framework demonstrates acceptable resilience even when forced to use a random, non-ideal reference sample, relying on random selection introduces high variance and unpredictability. The centroid-based approach is highly recommended as it acts as an "insurance policy," systematically mitigating the risk of catastrophic failure and ensuring robust, reproducible segmentation performance across diverse domains.